%% file: main.tex
\documentclass[10pt,twocolumn,letterpaper]{article}

\usepackage{iccv}
\usepackage{times}
\usepackage{epsfig}
\usepackage{graphicx}
\usepackage{amsmath}
\usepackage{amssymb}
\usepackage{multirow}
\usepackage{bm}
\usepackage[ruled,vlined]{algorithm2e}

\usepackage{booktabs}

\usepackage[pagebackref=true,breaklinks=true,letterpaper=true,colorlinks,bookmarks=false]{hyperref}

\newcommand{\scal}[1]{\mathit{#1}}
\newcommand{\vect}[1]{\mathbf{#1}}
\newcommand{\matr}[1]{\mathbf{#1}}
\newcommand{\const}[1]{\mathit{#1}}
\newcommand{\set}[1]{\mathcal{#1}}

\iccvfinalcopy 

\def\iccvPaperID{7757} 

\begin{document}

\title{Label-Efficient Point Cloud Semantic Segmentation: \\ An Active Learning Approach}

\author{Xian Shi$^1$, Xun Xu$^2$, Ke Chen$^1$, Lile Cai$^2$, Chuan Sheng Foo$^2$, Kui Jia$^1$\\
$^1$South China University of Technology\\
$^2$I2R, A*STAR\\
{\tt\small ausxian@mail.scut.edu.cn}, {\tt\small alex.xun.xu@gmail.com}, {\tt\small chenk@scut.edu.cn},\\ {\tt\small \{caill,3Q foo\_chuan\_sheng\}@i2r.a-star.edu.sg}, {\tt\small kuijia@scut.edu.cn}
}

\maketitle
\begin{abstract}
Deep learning models are the state-of-the-art methods for semantic point cloud segmentation, the success of which relies on the availability of large-scale annotated datasets. However, it can be extremely time-consuming and prohibitively expensive to compile such datasets. In this work, we propose an active learning approach to maximize model performance given limited annotation budgets. We investigate the appropriate sample granularity for active selection under realistic annotation cost measurement (clicks), and demonstrate that super-point based selection allows for more efficient usage of the limited budget compared to point-level and instance-level selection. We further exploit local consistency constraints to boost the performance of the super-point based approach. We evaluate our methods on two benchmarking datasets (ShapeNet and S3DIS) and the results demonstrate that active learning is an effective strategy to address the high annotation costs in semantic point cloud segmentation.
\end{abstract}

\section{Introduction}
\input{Introduction}

\section{Related Work}
\input{Relatedwork}

\section{Methodology}
\input{Methodology}

\section{Experiment}
\input{Experiment}

\section{Conclusion}
In this work we explored how active learning can help reduce annotation costs in the context of 3D point cloud segmentation. We explored several specific adaptations for the task, introducing super-point based labelling, a shape-based diversity measure, and a realistic annotation cost measurement. We also introduce a loss to encourage consistency of predictions within a super-point. Our experiments on two state-of-the-art 3D point cloud datasets showed that we can obtain more than $95\%$ of the performance of the fully supervised method using only 200k clicks, equivalent to $0.8\%$ clicks one could make exhaustively. Our method also outperforms all existing works on point cloud segmentation with less labeled data. We also observed that random selection provides an unusually strong baseline in the context of 3D point cloud segmentation, even outperforming traditional diversity and uncertainty selection that has performed well on other tasks. Our work shows that active learning is a feasible and competitive approach for reducing annotation costs for 3D point cloud segmentation, and provides a foundation for future research in this direction.

{\small
\bibliographystyle{ieee_fullname}
\bibliography{reference}
}

\clearpage
\textbf{\Large Appendix}
\appendix

\section{Active Learning Experiments on PartNet Dataset} \label{sec1}
{We evaluate active learning for part segmentation on PartNet\cite{mo2018partnet} dataset. PartNet is a large-scale fine-grained point cloud dataset, consisting of 26671 shapes from 24 unique categories. It involves three segmentation levels as coarse-, middle- and fine-grained and we evaluate our algorithms at level 1. For the sake of fair comparison, we sample 2048 points from each shape with furthest point sampling and cluster 500 super-points within each shape. The experiment results with label budget as 10k/20k/50k/100k/300k are presented presented in Tab. \ref{PartNet evaluation}. Before active selection loop, the first labeled pool is randomly initialized with 5k budget. It can be found that the segmentation mIoU under our method is consistently better than Random Selection, reflecting the universality of Active Selection.}

\section{Active Selection Parameters} \label{sec2}
{To obtain the optimal active selection parameters, we carried out a series of $\beta$ and $\delta$ comparison experiments under 200k label budget in Tab. \ref{Active Selection Parameters}. It shows that when $\beta=0.25$ and $\delta=1.0$, the experiment obtains the best result, reaching a {$79.9\%$} mIoU. In addition, it can be also found that mIoU results are more sensitive to $\delta$ than $\beta$. This explains that it 
necessary to balance the annotation budget among shapes}

\section{NuclearLoss Weights} \label{sec3}
{We also conducted comparison experiments on the nuclear loss weights $\lambda_{nc}$, with the best 200k active selected labels. The results displayed in Tab. \ref{NuclearLoss Weights}. prove that $\lambda_{nc}=1.0$ may be closest to the optimal value, obtaining a {$79.9\%$} mIoU. It led to a {$0.5\%$} mIoU improvement compared to the result without nuclear loss. }

\begin{table}[!t]
	\caption{Segmentation evaluation (mIoU$\%$) on PartNet}
	\label{PartNet evaluation}
	\centering
	\resizebox{1.0\linewidth}{!}{
	\begin{tabular}{c c c c c c c}
		\toprule
		\multicolumn{2}{c}{Dataset} & \multicolumn{5}{c}{PartNet} \\
		\hline
		\multicolumn{2}{c}{Full-Supervised} & \multicolumn{5}{c}{65.6} \\
		\hline		
		\multicolumn{2}{c}{Budget (\# clicks)} & 10k & 20k & 50k & 100k & 300  \\
		\hline			
		\multirow{2}{*} {Super-Point} & Random & $38.9$ & $42.6$ & $46.1$ & $51.2$ & $54.2$  \\
		& Active & \bm{$40.9$} & \bm{$43.1$} & \bm{$47.0$} & \bm{$51.4$} & \bm{$54.7$} \\
		\bottomrule
	\end{tabular}
	}
\end{table}

\begin{table}[!t]
	\caption{Active Selection Parameters Comparison on ShapeNet}
	\label{Active Selection Parameters}
	\centering
	\begin{tabular}{c c | c c}
		\toprule
		$\beta(\delta=1.0)$ &   mIoU$(\%)$  & $\delta(\beta=0.25)$ &  mIoU$(\%)$ \\
		\hline
		     $0.0$       &     $78.9$        &  $0.0$           &    $76.8$      \\
		     $0.25$       &     \bm{$79.9$}        &  $0.5$           &    $78.9$      \\
		     $0.5$       &     $79.2$        &  $1.0$           &    \bm{$79.9$}      \\
		     $0.75$       &     $78.8$        &  $2.0$           &    $77.6$      \\
		     $1.0$       &     $78.2$        &  $5.0$           &    $76.6$      \\
		\bottomrule
	\end{tabular}
\end{table}

\begin{table}[!t]
	\caption{NuclearLoss Weights Comparison on ShapeNet}
	\label{NuclearLoss Weights}
	\centering
	\begin{tabular}{c | c c c c c}
		\toprule
		$\lambda_{nc}$ & $0.0$ & $0.5$ & $1.0$ & $2.0$ & $5.0$  \\
		\hline
		mIoU$(\%)$ & $79.4$ & $79.7$ & \bm{$79.9$} & $78.5$ & $77.8$  \\
		\bottomrule
	\end{tabular}
\end{table}

\section{Additional Qualitative Results for Super-Point} \label{sec4}
\begin{figure*}[!t]
\centering   
\includegraphics[scale=0.6]{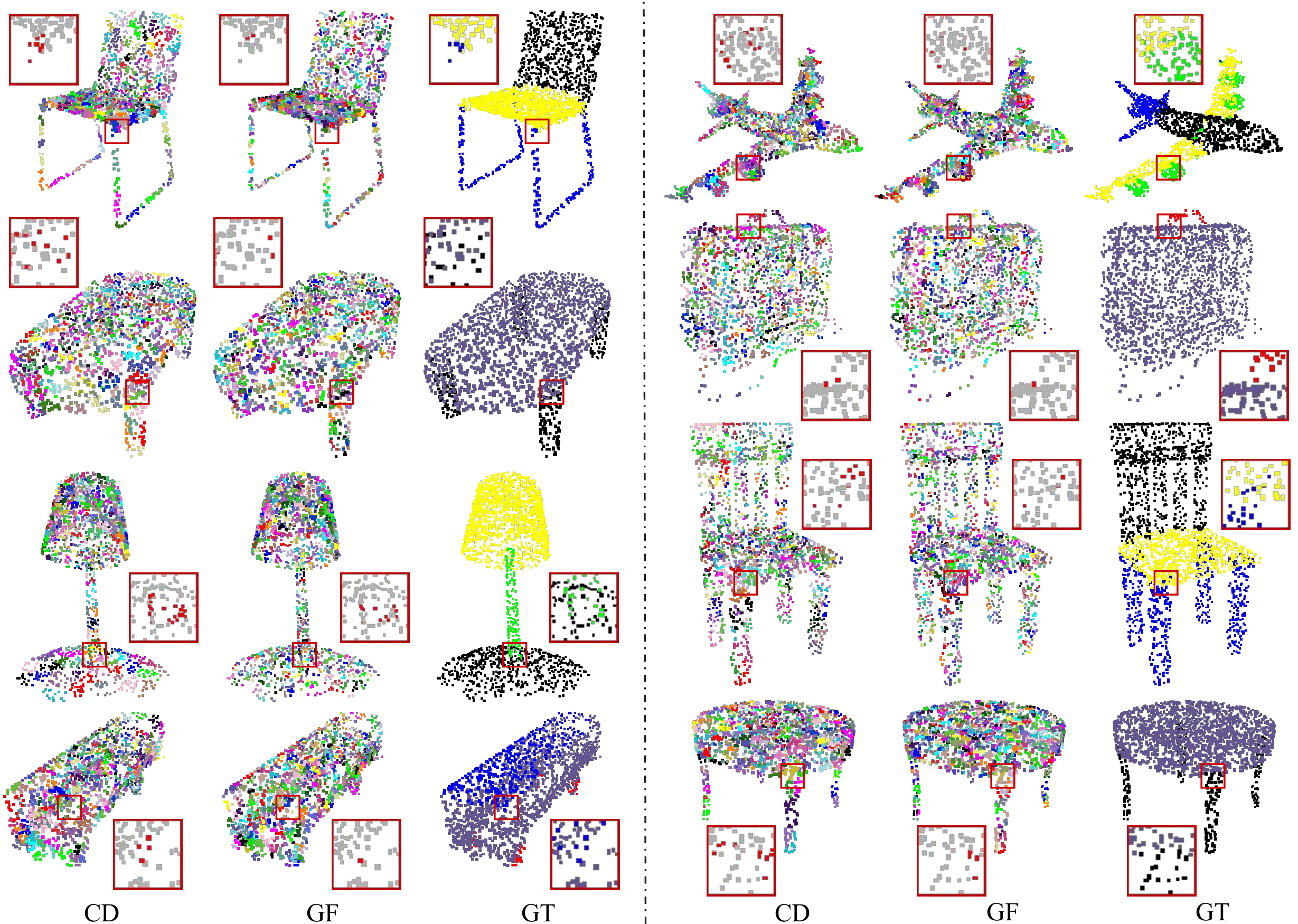}
\caption{Super-points clustered with spatial coordinate (CD) v.s. with Geometric Features (GF) v.s. Ground Truth (GT). Each shape is clustered into 500 super-points and different clusters are shown in different colors. The red points in zoomed-in boxes of the first and second columns represent the wrongly assigned points caused by super-points spanning across semantic boundary.}
\label{super-point generation}
\end{figure*}

We show more examples, on ShapeNet, clustered with spatial coordinate only (CD) v.s. with additional Geometric features(GF) and the corresponding semantic ground truth(GT) in Fig.~\ref{super-point generation}. For each shape in training set, We cluster it into 500 super-points without overlap. Since we assign each super-point with the majority label of the points it contains, those super-points that span across semantic boundaries lead to a small number of wrongly labeled points, which are marked in red in the figure. It can be seen from the figure that the super-points generated by geometric features better resembles the semantic boundaries, resulting in fewer red points.

\section{Additional Qualitative Results for Active Learning} \label{sec5}
\begin{figure*}[!t]
\centering   
\includegraphics[scale=0.8]{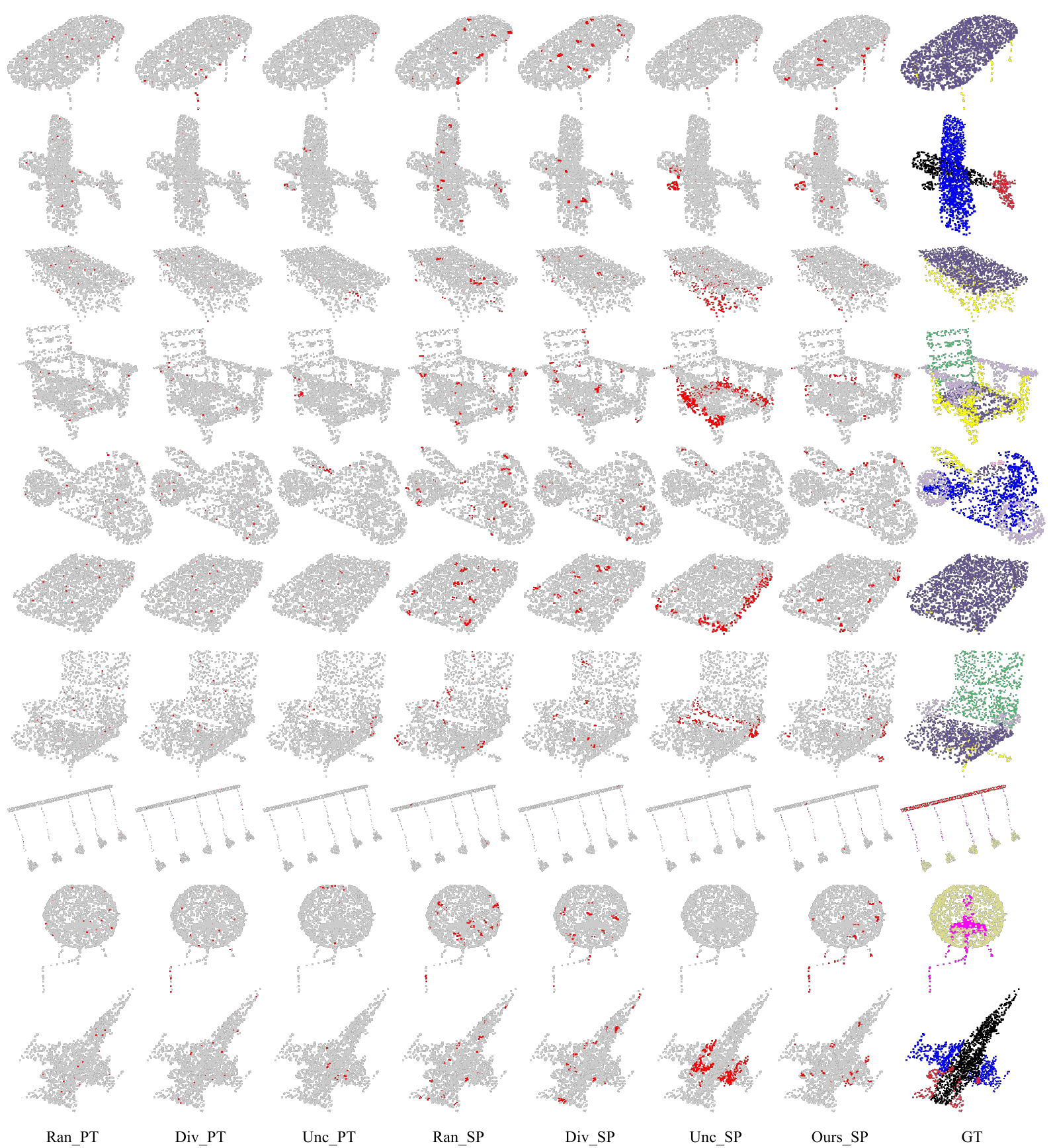}
\caption{We visualize selected points/super-points under different active selection strategies in $200k$ labeling budget on ShapeNet data. Ran\_PT, Div\_PT and Unc\_PT respectively represent random, diversity and uncertainty selection on point level data. And Ran\_SP, Div\_SP, Unc\_SP and Ours\_SP are the results of super-point level data selected with random, diversity, uncertainty and our method. The selected units are highlighted in red. In the final column objects semantic segmentation ground truth (GT) are shown.}
\label{shapenet active selection}
\end{figure*}

\begin{figure*}[!t]
\centering   
\includegraphics[scale=0.8]{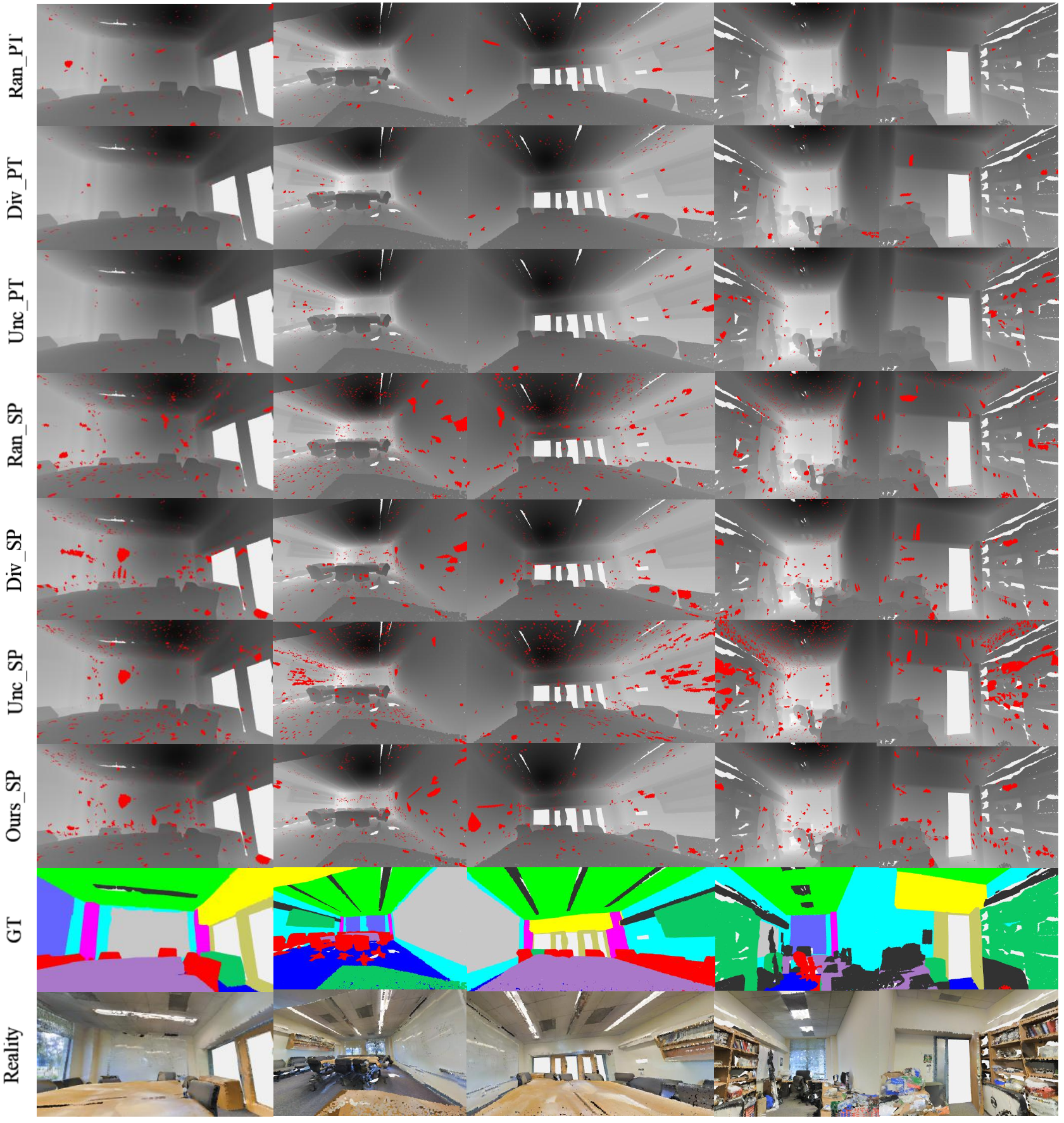}
\caption{We visualize selected points/super-points under different active selection strategies with $1m$ labeling budget on S3DIS data. Ran\_PT, Div\_PT and Unc\_PT respectively represent random, diversity and uncertainty selection on point level data. And Ran\_SP, Div\_SP, Unc\_SP and Ours\_SP are the results of super-point level data selected with random, diversity, uncertainty and our method. The selected units are highlighted in red. The last two lines show the semantic ground truth (GT) and the real scan RGB images.}
\label{s3dis active selection}
\end{figure*}

We present more qualitative active selection results for ShapeNet and S3DIS in Fig.~\ref{shapenet active selection} and Fig.~\ref{s3dis active selection}. We fix total annotation budget for ShapeNet and S3DIS data respectively at 200K and 1m. Good active learning algorithms are expected to select a diverse while informative set of points/super-points. 

In Fig.~\ref{shapenet active selection}, the 1st to the 6th columns show the points/super-points selected by random (Ran\_PT/Ran\_SP), diversity (Div\_PT/Div\_SP) and Uncertainty (Unc\_PT/Unc\_SP) strategies on ShapeNet. The 7th column displays our method conducted on super-points(Ours\_SP) and the last column is ground-truth (GT). The selected points/super-points are marked in red. It can be seen from the picture that, compared with individual points, super-points have a strong covering ability. A small number of labeled super-points can cover a substantial surface of the object, which is conducive to the efficient learning of a network. In the meantime, we can find that Diversity strategy (Div\_PT/Div\_SP) tends to select the points/super-points of scattered distribution, but ignores the complex areas of the object. Uncertainty (Unc\_PT/Unc\_SP) can find more ambiguous areas (such as aircraft wheel) in the object, but it is more likely to spend a large amount of annotation budget in one position, resulting in the lack of diversity of selected points. Our method takes into account the advantages of both methods and combines shape diversity to select a set of super-points with high complexity and diversity. 

We also visualize the super-points selected for S3DIS dataset. Since we divide the S3DIS data into cubicles during training, for visualization in the whole room we stitch cubicles together to recover the labeled super-points in each room. Selected super-points under different strategies are visualized in Fig.~\ref{s3dis active selection}. It demonstrates that under random selection, the selected points are more likely to fall on the structure covering larger area such as ceiling and floor due to the imbalanced distribution of semantic classes. In contrast, out method does better in selecting more super-points to label from minority categories, e.g. the whiteboard in the 1st column of Fig.~\ref{s3dis active selection}.

\section{Qualitative Results for Segmentation} \label{sec6}
\begin{figure*}[!t]
\centering   
\includegraphics[scale=0.9]{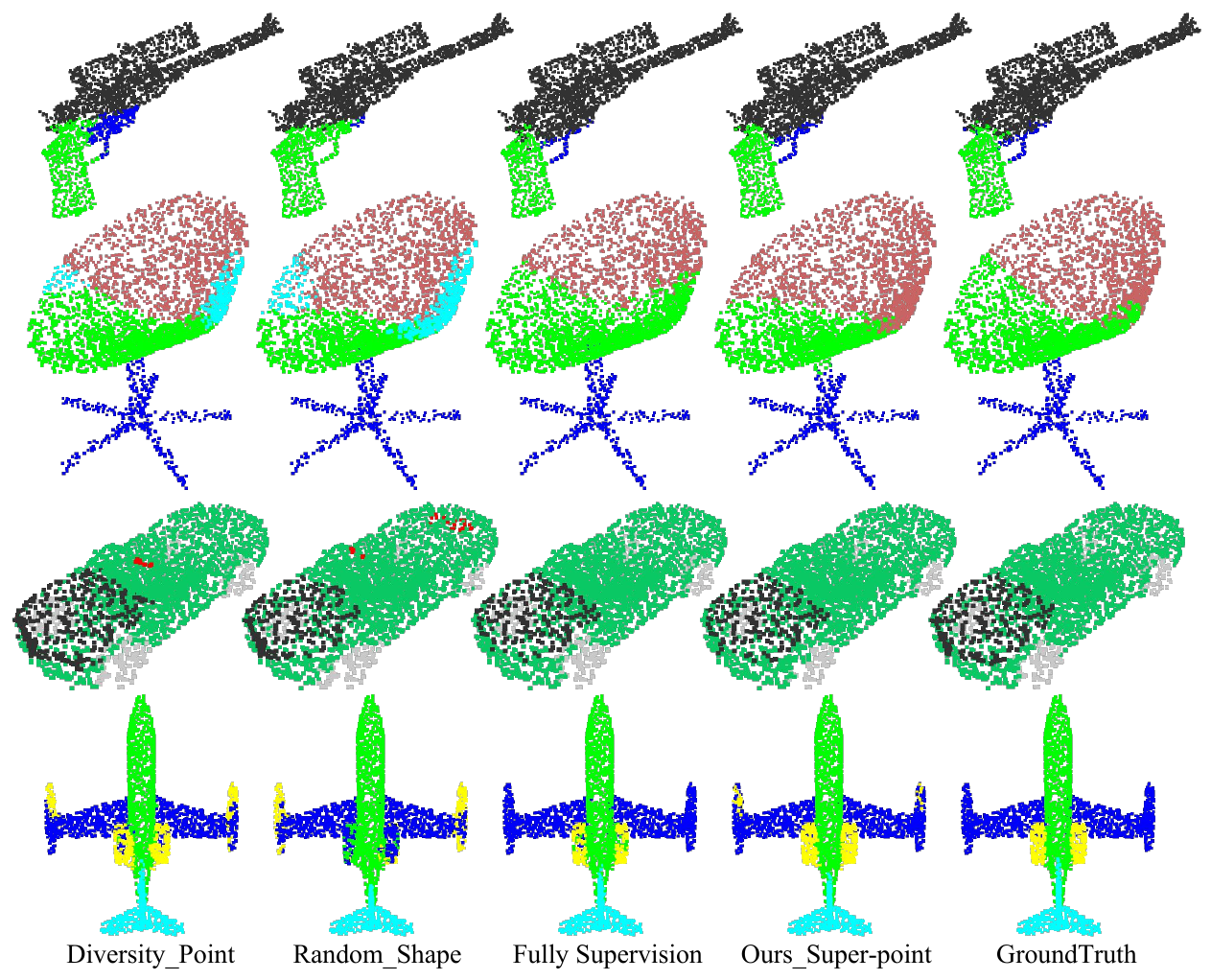}
\caption{We present part segmentation results on ShapeNet comparing different methods as Diversity Point(diversity selection on point level), Random Shape(random selection on shape level), Fully Supervision, Ours Super-point(our methods on super-point level) and Ground-Truth.}
\label{shapenet performance}
\end{figure*}

\begin{figure*}[!t]
\centering   
\includegraphics[scale=0.6]{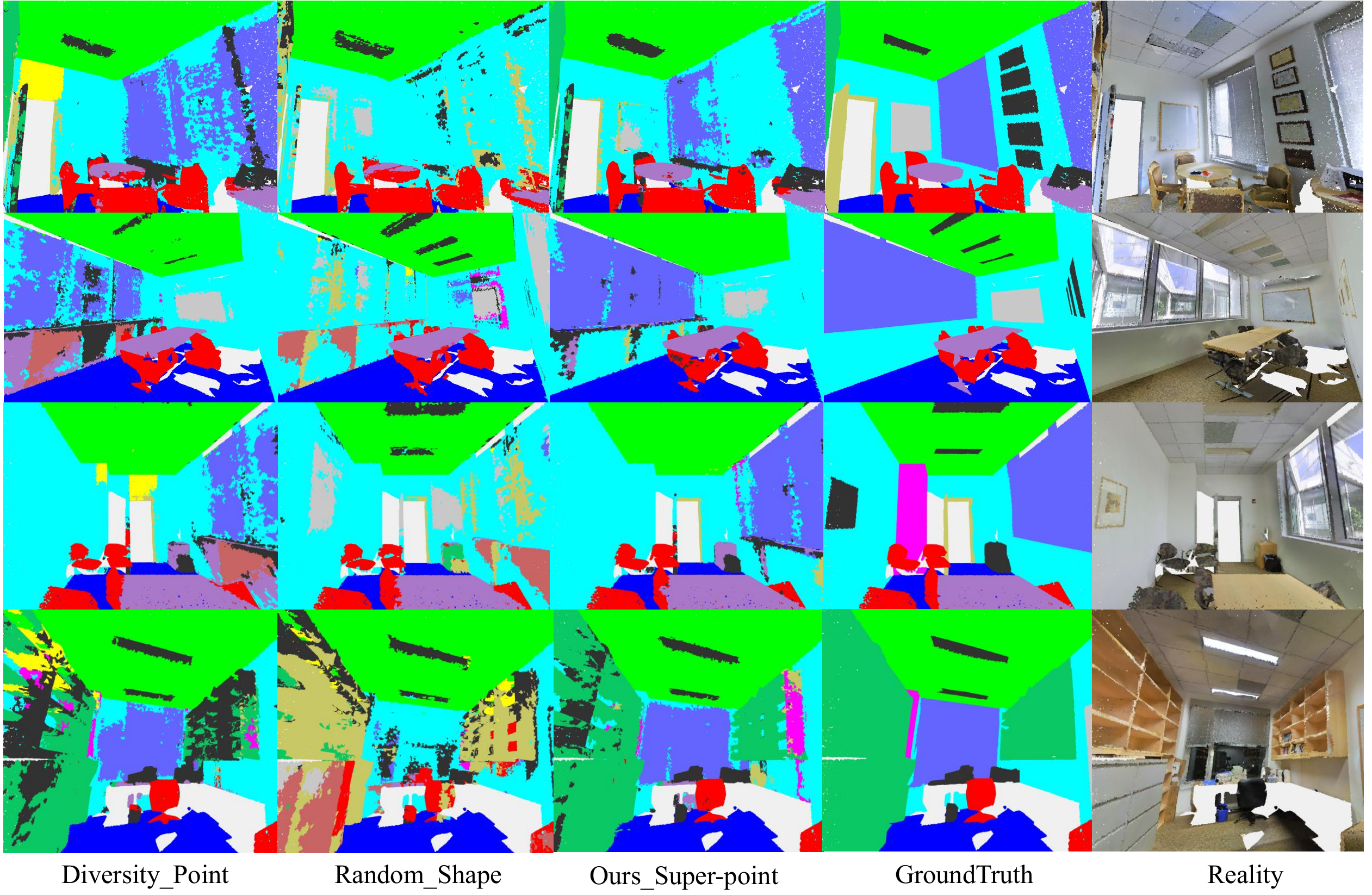}
\caption{We present semantic segmentation results on S3DIS among different methods as Diversity\_Point(diversity selection on point level), Random\_Shape(random selection on shape level) and Ours\_Super-point(our methods on super-point level). In the last two columns, segmentation ground-truth and real-scan RGB images are presented.}
\label{s3dis performance}
\end{figure*}

We present qualitative segmentation results with model trained on selected points/super-points on ShapeNet and S3DIS in Fig.~\ref{shapenet performance} and Fig.~\ref{s3dis performance}. For ShapeNet, we fix the labeling budget at 200k and present the methods with best performance at different levels. At point level the best active selection strategy is diversity (Diversity Point). Random selection performs best at shape level data (Random Shape). Our method does the best at super-point level (Ours Super-point). We also show the Fully Supervision results and Ground-Truth in Fig.~\ref{shapenet performance}. It can be observed that our method outperforms all alternative active learning methods, and the segmentation results are closer to that of fully supervised results. The segmentation results in details of the `\textit{gun}' are even better than the results of full supervision.

For S3DIS, under 1m label budget, we present the best segmentation performance at point (Diversity Point), shape (Random Shape) and super-point (Ours Super-point) levels, as wekll. The segmentation ground-truth and real-scan RGB images are shown in the last two columns. It can be clearly seen in the figure that our method has obvious advantages over other methods in `window' and `table'. These results demonstrate the superiority of our methods in semantic segmentation task.

\begin{figure*}[!t]
\centering   
\includegraphics[scale=0.5]{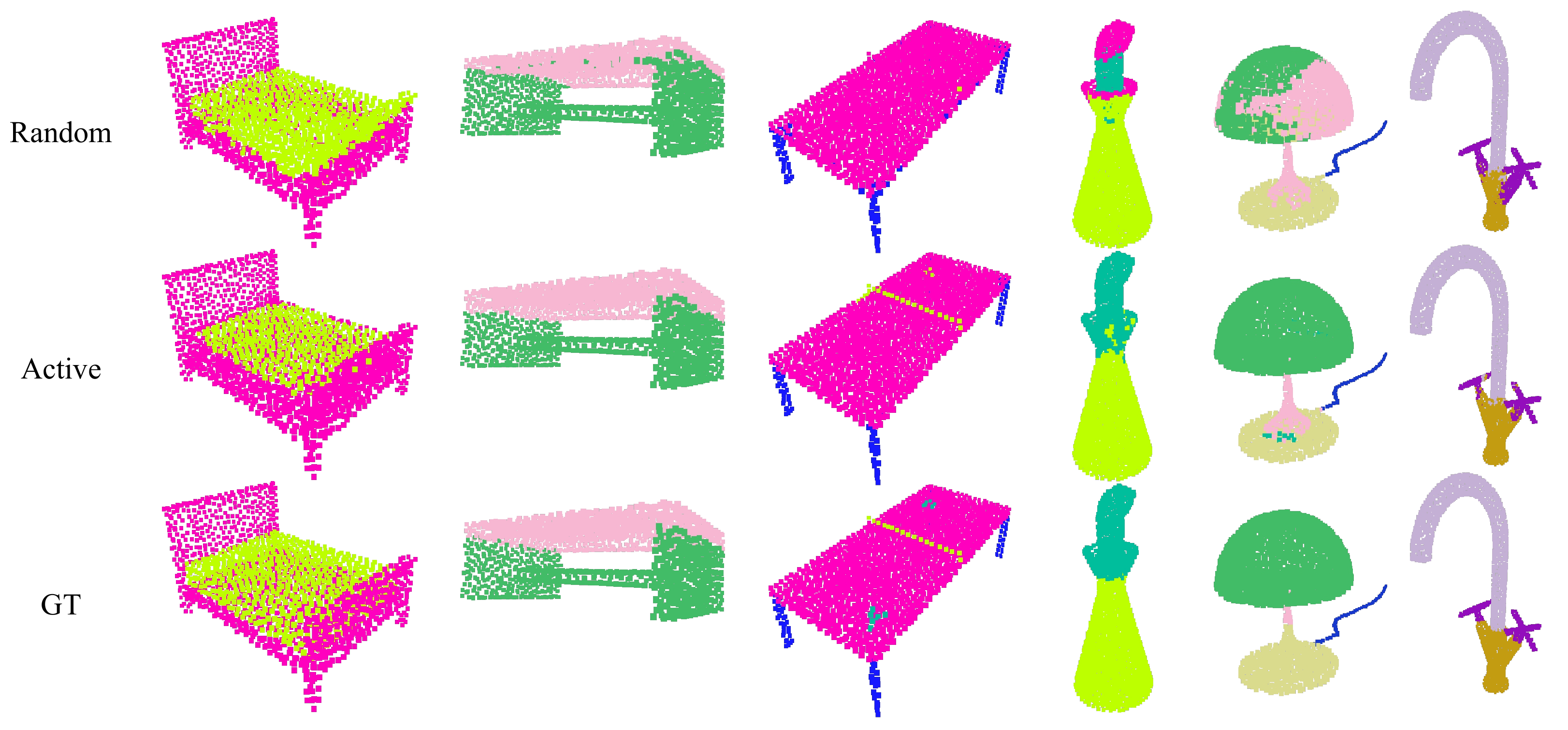}
\caption{We present PartNet segmentation results under Random/Active Selection and the corresponding ground truth(GT). }
\label{partnet performance}
\end{figure*}

{In Fig.~\ref{partnet performance}, we present PartNet fine-grained segmentation results with 300k annotating budget under Random/Active Selection. Compared with Random Results, Active Segmentation is more reasonable and closer to the third line Ground Truth, especially in the details such as the table net and faucet switch.}

\end{document}

%% file: Introduction.tex
\begin{figure*}[ht]  
    \centering
    \includegraphics[scale=0.5]{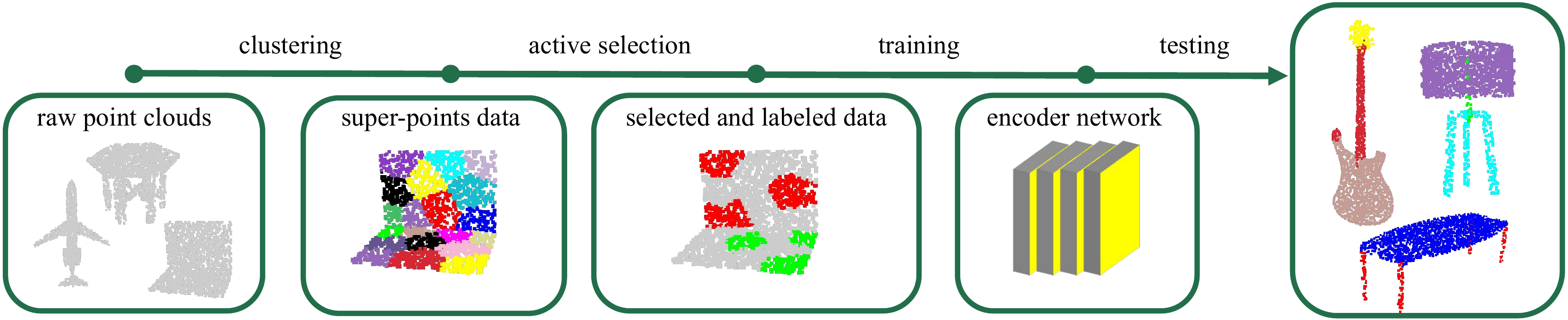}
    \caption{The overview of active learning for 3D point cloud segmentation.}
    \vspace{-0.5cm}
    \label{fig:my_label}
\end{figure*}

Semantic point cloud segmentation has attracted much interest recently due to its wide application in autonomous navigation, and many works have focused on the design of model architectures for this task \cite{choy20194d,wang2019dynamic,thomas2019kpconv}. Training these models, however, demands large amounts of annotated data, which can be extremely time-consuming and expensive to compile. To reduce these annotation costs, recent work has focused on label-efficient learning strategies for point cloud segmentation, including semi-supervised learning and unsupervised learning \cite{xu2020weakly,wei2020multi}. 

Besides these learning strategies, active learning (AL) is another traditional approach that can address these high annotation costs.
The goal of AL is to obtain similarly high performance with a fraction of the annotation costs, by choosing an informative and diverse subset of the data to label. Due to the data-hungry nature of deep learning (DL), there has been a resurgence of interest in AL in recent years \cite{ren2020survey}, and it has been investigated extensively for various DL applications, for instance, visual data processing and natural language processing (as reviewed in \cite{ren2020survey}). However, to the best of our knowledge, the use of active learning for 3D point cloud segmentation has not been explored.

In this work, we study the application of active learning to 3D point cloud segmentation, and explore specific adaptations for this task. More concretely, we consider the key aspects of \emph{sample selection} and \emph{annotation cost metric}. We discuss each of these below:

\emph{Sample selection}. We explore both sample selection granularity and acquisition functions. A key consideration when applying AL for point cloud data is the granularity of sample selection. Point-level selection will dramatically increase annotation costs, while instance-level selection can cause the annotation budget to be wasted on labeling redundant area within a shape. Instead, we propose to perform active selection based on super-points, which are generated by grouping geometrically similar points. Each super-point contain points that are most likely from the same category, and thus can be conveniently annotated by only one class label. We also introduce a shape diversity term in the acquisition function for sample selection.

\emph{Annotation cost metric}. To fairly evaluate the effectiveness of different selection strategies, we propose to measure the annotation cost by clicks, instead of the traditionally used point-based measurement. This is motivated by the fact that the actual annotation process is based on the annotator manipulating mouse clicks on the labelling interface. Click-based measurement has been investigated for AL in 2D semantic segmentation \cite{mackowiak2018cereals,metabox2020}, yet remains unexplored for 3D point cloud segmentation. Our work fills this gap.

We conduct comprehensive studies on these important design choices for applying AL on 3D point cloud segmentation, evaluating sample selection at various granularities and different acquisition functions. We further propose a local consistency loss on model predictions to boost performance using the unlabeled data (that is yet to be labeled).

Our contributions can be summarized as follows:
\begin{itemize}
    \item To the best of our knowledge, this is the first work exploring active learning for 3D point cloud segmentation. Our work demonstrates AL is an effective label-efficient learning strategy for this task.
    \item We investigate the effectiveness of various selection granularities, comparing point-level, superpoint-level and instance (shape)-level selection schemes using realistic, click-based annotation costs. Our results show that superpoint-based selection is the most efficient way to utilize a limited annotation budget.
    \item We investigate the effectiveness of various acquisition functions to perform sample selection. Specifically, we introduce an additional shape diversity term to complement the commonly used feature diversity and uncertainty terms in the acquisition function.
    \item We further exploit local consistency constraints within each super-point to boost model performance.
\end{itemize}

%% file: Relatedwork.tex
\subsection{Active Learning}
\vspace{-0.1cm}
Active learning has been extensively studied to optimally exploit the annotation budget. It aims to select the most informative data samples from all unlabeled data for an oracle to annotate. Commonly adopted active learning approaches consider diversity \cite{2003Incorporating} and uncertainty \cite{Joshi2009Multi} as the clue to guide the selection. In specific, \cite{sener2017active} proposed to maximize the diversity of selected samples in feature space and formulated the problem as a K-center problem. Uncertainty is often measured by the entropy of classifier's posterior. A recent work adopted an adversarial training strategy to measure the uncertainty where samples that can be easily attacked should be labelled \cite{2018Adversarial}. In this work, we discover that the spatial adjacency property on point cloud is not considered in generic active learning research. Therefore, we propose to use the spatial property as additional property to guide the active selection. 

\vspace{-0.1cm}
\subsection{3D Point Cloud Analysis}
\vspace{-0.1cm}
3D point cloud can be collected from LiDAR sensor and has been applied in autonomous driving, robotics, etc. Deep learning on 3D point cloud is mostly focused on developing backbone networks, e.g. PointNet \cite{Qi2017}, PointNet++ \cite{qi2017pointnet++}, DGCNN \cite{wang2019dynamic}, etc. These methods were demonstrated on recognition \cite{wu20153d,yi2016scalable}, semantic segmentation \cite{yi2016scalable,mo2019partnet}, object detection \cite{qi2019deep}. Among these semantic segmentation has wide applications in autonomous driving and robotics and it is particularly difficult to label segmentation masks in 3D point cloud. Therefore, we choose to demonstrate active learning on 3D point cloud semantic segmentation tasks.

\vspace{-0.4cm}
\paragraph{Point Cloud Annotation}
Labeling 3D point cloud requires assigning labels to each individual point. For individual shape point clouds, \cite{yi2016scalable} proposed an active framework to iteratively generate and validate manual and automatic annotation, achieving a point cloud annotation procedure with both efficiency and accuracy. For large scale scene point cloud, \cite{armeni20163d} first parsed the whole raw point clouds into individual physical spaces, and then annotated each independent space in the way of detection. However, there is no standard annotation method and no agreed metric to quantify the cost associated with 3D point cloud annotation. 
{Due to the complex rotation and transformation, annotating samples in 3D space is challenging. With the help of virtual reality, \cite{8814115,6798885} reconstructed and annotated the point cloud in the virtual world, where objects could be easily translated and rotated. Many 3D point cloud annotating tasks are conducted on 2D display screens. \cite{8374583} selected and annotated some control points by clicking the mouse, and then the control labels would be propagated to other points according to the shortest path of the neighborhood graph. During labeling, to select a specific part from a whole sample, \cite{6798882} uses gesture sensors to slice and slide the point cloud until the rest is the desired set. In order to label scene point cloud quickly and accurately, \cite{DBLP:journals/corr/WongCM14} proposed a geometric and structural guided annotation model, which would be improved under manual correction. Despite so many research into annotation tool, there is no research into selectively annotating data samples which is the focus of this work. }

\vspace{-0.4cm}
\paragraph{Label-Efficient Learning}
Existing research often assume a large fully labelled dataset available which is sometimes too strong. It is known that labelling 3D point cloud is expensive and sometimes requires expertise \cite{xu2020weakly}. As a result, the community recently turned to exploiting partially labelled data \cite{xu2020weakly}, inexactly labelled data \cite{wei2020multi} and weakly labelled data \cite{qin2020weakly} for the sake of saving annotation cost. In addition, there are several ways to improve network performance with insufficient labeled data. Capsule network \cite{2018On, zhao20193d} can maximally reserve valuable information and infer possible variables with less training data. \cite{9009542} combines three unsupervised learning methods including clustering, reconstruction and self-supervised classification to improve network performance on less labelled data.  Pre-training \cite{xie2020pointcontrast} is also proved to be able to deal with low capacity network when there are few labeled samples. Nevertheless, there is still no rigorous study of whether or how much annotation is saved by various annotation schemes. In this work, we provide a fair comparison of annotation cost and propose several ways to achieve better performance with less labeled data.

%% file: Methodology.tex
We first give an overview of the proposed active learning framework for point cloud segmentation. We then detail each component, including super-point generation, selection strategies and local consistency constraints.

\subsection{Overall Framework}
Given a set of unlabelled point cloud samples, our method first groups individual points into super-points. We then select a small subset of samples to label based on the common selection criteria to maximize diversity and uncertainty of the selected samples. The model is then re-trained with all the samples labeled so far. The select-train process is iterated until the annotation budget is exhausted.

\subsection{Super-point Generation}

\paragraph{Spectral Clustering} 
{
We employ spectral clustering for super-point generation. Given a point cloud sample with $N$ points $\set{X}_i=\{\vect{p}_{ij}\}_{j=1\cdots N}$, an affinity matrix $\matr{A}$ is first computed characterizing the adjacency of data points. All points are then grouped into $K$ non-overlapping clusters by normalized cut algorithm \cite{2000Normalized}. To construct the affinity matrix, one could build a 
$k$ nearest neighbor graph based on the Euclidean distance defined over spatial coordinates and RGB values \cite{xu2020weakly}. However, we argue that Euclidean distance alone is still prone to non-uniform point distribution and complex local geometric structures, thus making erroneous links. To remedy this issue, we propose to exploit the local geometric feature to take into account the 
local geometric structure.
}

\begin{figure}[!t]
\centering 
\includegraphics[scale=0.5]{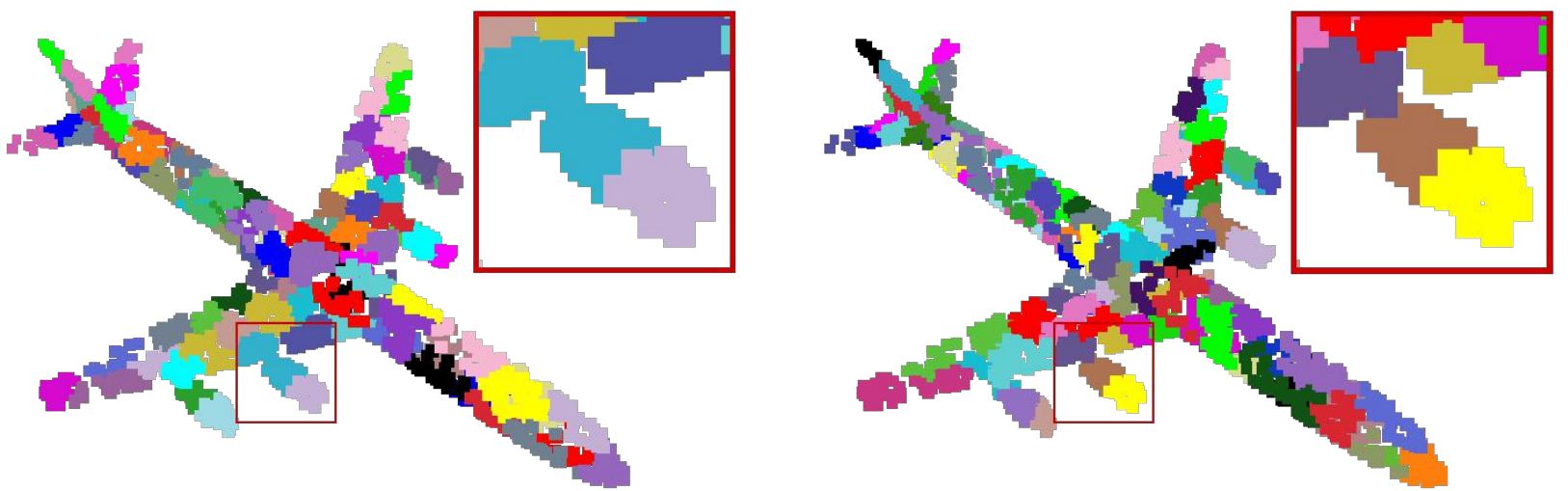}
\caption{Super-points clustered with spatial coordinates v.s. with additional geometric features. Color indicates different super-points.}
\label{fig:SuperPoint}
\end{figure}

\paragraph{Geometric Features} 
{To characterize the local shape of the region around a point, we compute three geometric features \cite{J2011DIMENSIONALITY} by looking into the neighborhood. For a point $\vect{p}=(x \; y \; z)$, we denote its $k$ nearest neighbors as $\set{NN}_k(p)$. Given $\matr{M}=(\vect{p}_1-\vect{p}, \quad...\quad ,\vect{p}_n-\vect{p})^T$, where $\vect{p}_1 \;...\; \vect{p}_n \in \set{NN}_k(\vect{p})$, the covariance matrix is defined by $\matr{C}=\frac{1}{n} \matr{M}^T\matr{M}$. Since $\matr{C}$ is a symmetric positive semi-definite matrix, its three eigenvalues exist and are non-negative. Without loss of generality, we define eigenvalues as $\scal{\lambda}_1 \geq \scal{\lambda}_2 \geq \scal{\lambda}_3 \neq 0$. The linearity $(\scal{f}_1)$, planarity $(\scal{f}_2)$ and scatterness $(\scal{f}_3)$ derived from the eigenvalues are described as,
\begin{equation}
f_1 = \frac{\lambda_1 - \lambda_2}{\lambda_1},\quad
f_2 = \frac{\lambda_2 - \lambda_3}{\lambda_1},\quad
f_3 = \frac{\lambda_3}{\lambda_1} 
\label{geo_fea}
\end{equation}
We define geometric feature vector as $\vect{g}=\left \{ \scal{f}_1,\scal{f}_2,\scal{f}_3 \right \}$ which derives a geometric feature distance between individual points as,
\begin{equation}
    d_g(\vect{p}_i,\vect{p}_j) = \left \| \vect{g}_i - \vect{g}_j \right \|_2
\end{equation}}
{The final affinity matrix $\matr{A}$ combines spatial distance and geometric feature distance as 
\begin{equation}
a_{ij} = \exp(-\frac{||\vect{p}_i-\vect{p}_j||^2}{2\sigma_s^2}) + \gamma \exp(-\frac{||\vect{g}_i-\vect{g}_j||^2}{2\sigma_g^2})
\end{equation}
where $\sigma_s$ and $\sigma_g$ are bandwidth parameters. Each resultant cluster $\set{C}$ constitutes a super-point $\set{S}=\{\vect{p}_i\}_{i\in\set{C}}$. An example comparing super-points generated with spatial coordinate alone and with additional geometric features are shown in Fig.~\ref{fig:SuperPoint}. Obviously, the latter one produces boundary between `engine' and `wing' that better aligns with the ground-truth semantic boundary.}

\subsection{Selection Strategies}
{
In this section, we first present the proposed annotation cost measurement scheme. We then describe the details of the acquisition function employed in this work and the proposed shape diversity term to complement existing feature diversity and uncertainty terms.
}
\paragraph{Annotation Cost Measurement}

Previous works for label-efficient learning in point cloud segmentation usually measure the annotation cost by the number of labeled shapes \cite{li2018so,xie2020pointcontrast} or the number of labeled points \cite{xu2020weakly}. We argue that neither method can reliably measure the real annotation cost as the efforts for labeling each shape can vary significantly across samples and annotation is not performed on a point basis. Motivated by the fact that the actual annotation process is based on the annotator manipulating mouse clicks on the labelling interface, we propose to measure the annotation cost by clicks. Under such measurement scheme, the cost for various selection granularies is computed as follows.\\
\textbf{Point-level selection}: under such a selection scheme, each point is assigned a class label and the number of involved clicks is equal to the number of points.\\
\textbf{Superpoint-level selection}: each super-point is assigned the class label of the majority points within it, and thus each super-point only costs one click to label.\\
\textbf{Shape-level selection}: we estimate the number of clicks used to annotate a shape based on the existing annotation procedure employed in each dataset.

\paragraph{Acquisition Function}

We first formally define the active point cloud annotation task. We denote all point cloud samples as $\set{X}=\{\set{X}_i\}$, e.g. airplane, car, etc. in ShapeNet dataset and cubicles in a room for S3DIS dataset, where $\set{X}_i=\{\set{S}_{j}\}$. Given limited labeling budget $B$, counted as the number of clicks in this context, we aim to select a subset of points, $\set{X}^L_i\subset \set{X}_i$, to annotate with $\set{Y}^L_i$; the selected subset is chosen by maximizing a specified acquisition function.
The annotation is usually provided in multiple batches and the active selection will repeat iteratively.
The core of active learning research is on designing the acquisition function to determine which data points to select for labeling. Feature diversity \cite{sener2017active} and uncertainty maximization \cite{Joshi2009Multi} are widely adopted acquisition functions that have performed well for classification and image segmentation. Nevertheless, we found both strategies locate many selected points on a single shape, which is not conducive to the generalization ability of network. To address this issue, we propose shape level diversity to balance the spread out of annotation over shapes, and combine it with feature diversity and uncertainty for our acquisition function ($AL_{score}$),
\begin{equation}
AL_{score} = (1 - \beta) D_{score} + \beta E_{score} + \delta S_{score}
\label{active learning score}
\end{equation}
where $\const{\beta}$ and $\const{\delta}$ are hyper-parameters. Our active selection follows a greedy algorithm by iteratively selecting the super-point with highest score, as summarized in Algo. \ref{AL algorithm}. We describe the individual terms in the following.

\paragraph{Feature Diversity ($D_{score}$)} emphasizes the scatter of distribution in feature space. \cite{sener2017active} proposed to maximize diversity measure as the objective for active learning and approximated with a greedy algorithm which only depends on an unary diversity score.
For a point $\vect{p}$, we denote $\varphi(\vect{p})$ as its high dimensional feature from network. Its diversity score is measured by $L2$ distance as
\begin{equation}
D_{score}(\vect{p}) = \min _{\vect{p}_j \in \set{X}_l} \left \| \varphi(\vect{p}) - \varphi(\vect{p}_j) \right \|_2
\label{points diversity}
\end{equation}
To extend diversity score to super-points $\set{S}$, we estimate a super-point feature by average pooling over all points it contains as
\begin{equation}
\varphi(\set{S}) = \frac{1}{\left | \set{S} \right |} \sum_{\vect{p} \in \set{S}}\varphi(\vect{p})
\label{diversity one}
\end{equation}
where $\left | \set{S} \right |$ represents the number of points in super-point $\set{S}$. Then for each super-point $\set{S}$ in unlabeled super-points pool $\set{X}_u$, its diversity score is formulated as,
\begin{equation}
D_{score}(\set{S}) = \min _{\set{S}_j \in \set{X}_l} \left \| \varphi(\set{S}) - \varphi(\set{S}_j) \right \|_2
\label{super-points diversity}
\end{equation}
\paragraph{Uncertainty ($E_{score}$)} measures network's confidence for the samples. Denoting $f(\vect{p}) \in R^k$ as the posterior prediction for a point $\vect{p}$, where $\const{K}$ is the number of predicted categories, it is common to adopt entropy as uncertainty score,
\begin{equation}
E(\vect{p}) = -\sum_{j\in k}f(\vect{p})^j\log(f(\vect{p})^j)
\label{entropy}
\end{equation}
For super-points level, we measure a super-point uncertainty by average pooling its points level uncertainty as,
\begin{equation}
E_{score}(\set{S}) = \frac{1}{\left | \set{S} \right |} \sum_{\vect{p} \in \set{S}}E(\vect{p})
\label{uncertainty}
\end{equation}
\paragraph{Shape Diversity ($S_{score}$)} As shapes in the dataset are of varying degrees of complexity, active selection may focus more on the shapes that are harder to learn. To avoid too many queried points falling on one sample, we propose shape diversity score to estimate shape level diversity in labeled data. For a shape point cloud $\set{X}$, we define its feature $\varphi(\set{X})$ as maxpooling of all points in the shape, the same as a super-point feature. Then we define shapes that have at least one annotated point as labeled shapes $\set{Y}_l$ and the rest as unlabeled shapes $\set{Y}_u$. The shape diversity score is measured in the same way as point diversity
\begin{equation}
S_{score}(\set{X}) = \min _{\set{X}_j \in \set{X}_l} \left \| \varphi(\set{X}) - \varphi(\set{X}_j) \right \|_2
\label{shapes diversity}
\end{equation}
It is worth mentioning that shape diversity score is a shape level value. All points in the same shape share one value.

\begin{algorithm}
\KwIn{Super-points Data $\set{X}$, Labels Budget $\const{B}$, Query Number $\const{n_{query}}$, Network Hyper-parameters $\set{\theta}$}
\KwOut{Labeled Points Set}
Randomly initialize super-points as labeled data $\set{X}_L$ and the rest as unlabeled pool $\set{X}_U$\\
\While{$\left | \set{X}_L \right | < \const{B}$}{
\# Train the network with current $\set{X}_L$\\
$\hat{\theta}=\arg\min_{\theta} l(\set{X}_L;\theta)$\\
\# Evaluate the super-points in $\set{X}_U$ and Compute the active scores $\set{S}$ by Eq.~(\ref{active learning score})\\
\For{$\set{X}_i \in \set{X}_U$}{$Score_{i}=AL_{score}(\set{X}_i)$}
\# Query super-point with the highest score\\ 
$\set{X_q}\leftarrow argsort(Score)$;
$\left | \set{X_q} \right | = \const{n_{query}}$\\
\# Update data pool\\
$\set{X}_L=\set{X}_L \cup \set{X_q} \quad s.t.\quad \set{X}_U=\set{X}_U \setminus \set{X_q}$}
\caption{Active Learning Algorithm}
\label{AL algorithm}
\end{algorithm}

\subsection{Geometric Consistency Constraint}
{As a large amount of data points are still unlabeled, we naturally consider taking advantage of semi and weakly supervised methods to extract extra information from the inherent characters of point cloud.}
{As we cluster super-points according to relative distances and local geometric shape in a point cloud, the points within a same super-point share high similarity, which inspires us to extend consistency loss on super-points. For a super-point $\set{S}$ covering $\const{N}$ points, the posterior prediction is defined as $f(\set{S}) \in R^{N \times C}$ where $C$ is the number of semantic categories. To enforce point level predictions within a super-point to be close to each other, we consider constraining the rank of $f(\set{S})$, which will reach the minimum if all predictions stay the same. Instead of directly minimizing the rank of $f(\set{S})$, we minimize nuclear norm in Eq.~(\ref{eq:NuclearNorm}), which is a convex approximate to matrix rank. 
\begin{equation}
l_{nc} = \frac{1}{N} tr(\sqrt{f(\set{S})^T f(\set{S})})
\label{eq:NuclearNorm}
\end{equation}
where $tr$ is the trace of a matrix. Finally our network loss comprises cross entropy loss for segmentation and nuclear loss. 
\begin{equation}
l_{total} = l_{seg} + \lambda_{nc} l_{nc}
\label{nuclear loss}
\end{equation}
}

%% file: Experiment.tex
\subsection{Dataset}
We use two 3D point cloud datasets to validate the proposed active learning framework.
\textbf{ShapeNet} \cite{yi2016scalable} consists of 16881 CAD shape models (12137, 1870 and 2874 for training, validating and testing respectively) from 16 categories with 50 part categories. For each shape model, 2048 points are sampled from mesh. 
\textbf{S3DIS} \cite{armeni20163d} was proposed for indoor scene analysis. It contains 3D point cloud with xyz coordinate and rgb information from 6 areas covering 271 rooms. We take Area 5 as testing set while the rest as training set. As each room is divided into 1*1 meter blocks with 4096 points, we conduct super-points clustering and active learning at each block in our experiments.

\subsection{Competing Methods}
 We compare our method with both single-point based labeling scheme and several variants of super-point based schemes described in the following.
 
 \noindent\textbf{Vanilla Point Based Scheme} 
 We evaluate random, feature diversity (Core-Set) \cite{sener2017active} and entropy \cite{Joshi2009Multi} based acquisition functions.

\noindent\textbf{Shape Based Scheme}
    To conduct active selection at shape level, the diversity and uncertainty of a shape are computed as the average diversity and uncertainty of all the points it contains. We evaluate the three acquisition functions as used in point-level selection.
    
\noindent\textbf{Super-point Based Scheme} On super-point data, we compare our active selection method with Random selection, feature diversity and entropy criterion. We further extend a data augmentation and our nuclear loss in the final results. 

\subsection{Configuration}
\paragraph{Data Augmentation}
{Data augmentation is a common way to avoid overfitting, especially when labels are insufficient. Although Gaussian perturbation is widely used in 2D images learning, it may not be an appropriate augmentation for point cloud on the semantic segmentation task. When the semantic boundaries are distinct in point cloud space, Gaussian perturbation is likely to destroy them as points may shift in arbitrary directions. To deal with this issue, we define a $3 \times 3$ deviation matrix as
\begin{equation}
\setlength{\abovedisplayskip}{0.1cm}
\setlength{\belowdisplayskip}{0.1cm}
\matr{T} = \matr{I} + \matr{G}
\label{deviation matrix}
\end{equation}
where $\matr{I}$ is a $3 \times 3$ identity matrix and $\matr{G}$ is a Gaussian matrix of the corresponding size. Then a shape point cloud $\matr{X} \in R^{N \times 3}$ can be augmented by
\begin{equation}
\setlength{\abovedisplayskip}{0.2cm}
\setlength{\belowdisplayskip}{0.1cm}
\widehat{\matr{X}} = \matr{X}\matr{T}
\label{augmentation}
\end{equation}
Through multiplying by the deviation matrix, a point cloud is transformed on rotation and scaling. As it is a linear transformation, the topology of the object is preserved, which holds the semantic boundaries. Besides, in consideration of the symmetry of most objects, we also apply the random mirroring on the corresponding axis.}
\vspace{-0.4cm}
\paragraph{Encoder Network} In view of the high training efficiency and outstanding semantic segmentation performance of DGCNN \cite{wang2019dynamic}, we choose it as our backbone network. For the efficiency of training, the network is trained 200 epochs on ShapeNet dataset and 100 epochs on S3DIS dataset. 
\vspace{-0.4cm}
\paragraph{Hyper-Parameters}
Before training, we offline compute the  geometric features, where the Knn Graph and neighbors is constructed by $\const{k}=10$. To balance the effects of geometric features in clustering, the hyper-parameter $\gamma$ is set as 0.1. In our experiments, we cluster each shape from ShapeNet into 500 super-points and S3DIS into 1000 super-points. In training section, loss weight $\lambda_{nc}$ is set as 1. During active selection, $\beta$ and $\delta$ in the formula is respectively assigned as 0.25 and 1.
\vspace{-0.4cm}
\paragraph{Evaluation Metric} For all testing data, we calculate the mean categories Intersect over Union (mIoU) as the evaluation metric. It is counted as the average of all category IoUs among the data.
\subsection{Active Learning for Semantic Segmentation}
\paragraph{Part Segmentation} Part segmentation attempts to assign part category label for the given 3D object. Each part category is considered as an independent class in the experiments. We evaluate active learning for part segmentation on ShapeNet with budgets as 10k/20k/50k/100k/200k respectively. As active learning iteratively picks samples for labeling, we randomly initialize the first labeled pool with 5k annotation budget, then in each iteration active learning selects candidate samples until the next budget set is reached. For example, with initialized 5k labeled data, network will be trained first. In the following selection phase, active learning will select another 5k samples to reach 10k budget. In the next iteration, additional 10k samples will be selected to reach the next 20k budget, until all budgets are exhausted. The results under different budgets compared with other competing approaches are presented in left of Tab.\ref{Table evaluation}. 
\vspace{-0.3cm}
\paragraph{Scenes Segmentation} Our approach is also applicable to semantic segmentation in scenes. We conduct semantic segmentation experiments on S3DIS dataset. We set training data label budget as 50k/100k/200k/500k/1m respectively. The first labeled pool is randomly initialized with 25k budget and latter active selection pattern is the same as that in part segmentation mentioned above. The results are presented in right of Tab.\ref{Table evaluation}.
\vspace{-0.3cm}
\paragraph{Analysis of Results}
Fully supervised baseline realizes the mIoUs of $82.3\%$ and $49.1\%$ respectively on ShapeNet dataset and S3DIS dataset. We make the following observations.
1) On ShapeNet data, our method on super-point data under 100k and 200k label budget achieved $79.9\%$ mIoU which is higher than $95\%$ relative performance of fully supervised approach ($78.4\%$). On S3DIS data, the results also show the superiority of our method, achieving $47.5\%$ mIoU with 1m label budget, ahead of $95\%$ fully supervised performance($46.6\%$).
2) Super-point method is consistently better than point level and shape level active learning at all labelling budgets, thus suggesting directly labeling on super-point is more label efficient. 
3) Our proposed AL algorithm is better than random, feature diversity and entropy criterion at all labelling budgets. This suggests the effectiveness of shape diversity for active selection.
4) When combined with consistency constraint (AL \& Const), we achieve the best performance of all competing methods. It is also consistently better than AL alone on both datasets.
5) Shape level annotation scheme is much worse than the other two under our cost counting scheme. It probably due to that annotation focused on limited number of shapes has negative impact on network learning. 
\begin{table*}[!t]
	\caption{Segmentation evaluation (mIoU$\%$) among different active learning strategies}
	\label{Table evaluation}
	\centering
		\resizebox{0.85\linewidth}{!}{
	\begin{tabular}{c c c c c c c | c c c c c}
		\toprule
		\multicolumn{2}{c}{Dataset} & \multicolumn{5}{c}{ShapeNet} & \multicolumn{5}{c}{S3DIS} \\
		\hline
		\multicolumn{2}{c}{Full-Supervised} & \multicolumn{5}{c}{$82.3$} & \multicolumn{5}{c}{$49.1$} \\
		\multicolumn{2}{c}{$95\%$ of Full-Performance} & \multicolumn{5}{c}{$82.3 \times 0.95 \approx 78.2$} & \multicolumn{5}{c}{$49.1 \times 0.95 \approx 46.6$} \\
		\hline		
		\multicolumn{2}{c}{Annotation Budget (\# clicks)} & $10k$ & $20k$ & $50k$ & $100k$ & $200k$ & $50k$ & $100k$ & $200k$ & $500k$ & $1m$ \\
		\hline			
		\multirow{3}{*} {Point Level} & Random & $63.3$ & $70.0$ & $73.0$ & $75.9$ & $76.0$ & $40.3$ & $41.1$ & $42.5$ & $45.0$ & $45.9$ \\
		& Core-set & $59.0$ & $66.2$ & $72.8$ & $76.0$ & $77.1$ & $39.0$ & $40.2$ & $42.0$ & $44.6$ & $46.6$  \\
		& Entropy & $59.5$ & $62.2$ & $63.1$ & $71.8$ & $73.4$ & $38.3$ & $41.3$ & $42.4$ & $43.3$ & $45.2$  \\
		\hline
		\multirow{3}{*} {Shape Level} & Random & $32.4$ & $33.1$ & $43.9$ & $45.8$ & $59.5$ & $13.6$ & $19.6$ & $22.2$ & $25.9$ & $31.2$ \\
		& Core-set & $27.7$ & $33.4$ & $34.5$ & $37.1$ & $44.7$ & $14.0$ & $19.7$ & $20.5$ & $22.8$ & $28.9$ \\
		& Entropy & $31.0$ & $32.0$ & $38.0$ & $44.1$ & $47.5$ & $13.7$ & $15.2$ & $17.0$ & $23.6$ & $24.3$ \\
		\hline
		\multirow{5}{*} {Super-point} & Random & $69.0$ & $70.8$ & $74.8$ & $76.1$ & $77.0$ & $41.5$ & $43.4$ & $44.3$ & $45.6$ & $46.4$ \\
		& Core-set & $65.2$ & $67.3$ & $70.1$ & $76.3$ & $77.4$ & $39.0$ & $40.6$ & $43.2$ & $44.5$ & $45.3$  \\
		& Entropy & $65.6$ & $68.4$ & $74.3$ & $75.7$ & $77.8$ & $40.1$ & $41.5$ & $43.7$ & $44.7$ & $45.2$  \\
		\cline{2-12}
		& Our AL & $70.5$ & $71.3$ & $76.4$ & $77.1$ & $79.4$ & $42.5$ & $43.8$ & $44.6$ & $45.9$ & $47.2$  \\
		& Our AL \& Const & \bm{$71.8$} & \bm{$74.1$} & \bm{$77.1$} & \bm{$78.4$} & \bm{$79.9$} & \bm{$42.8$} & \bm{$44.1$} & \bm{$45.9$} & \bm{$46.3$} & \bm{$47.5$} \\
		\bottomrule
	\end{tabular}
	}
\end{table*}



\subsection{Ablation Study}
To analyze the contribution of each individual component, we conduct ablation experiments on ShapeNet part segmentation and present the results in Tab.~\ref{Table Ablation}. The initial result is the performance of 10k labeled points by random selection. We make the following observations. 
1) Point level v.s. super-point (SP), we observe $1.1\%$ improvement from point data to vanilla super-point data.
2) Adding geometric feature (Geo) further improves $4.6\%$ mIoU compared with  super-points generated with point coordinate alone.
3) When AL considers shape diversity (Our AL), it does better than random selection with $1.5\%$ improvement. 
4) Nuclear loss for super-point consistency (Const) and Data Augmentation (DA) respectively boosts $0.9\%$ and $0.5\%$ performance. And the combination of both achieves $1.3\%$ improvement.

\begin{figure}[!t]
\centering   
\includegraphics[scale=0.5]{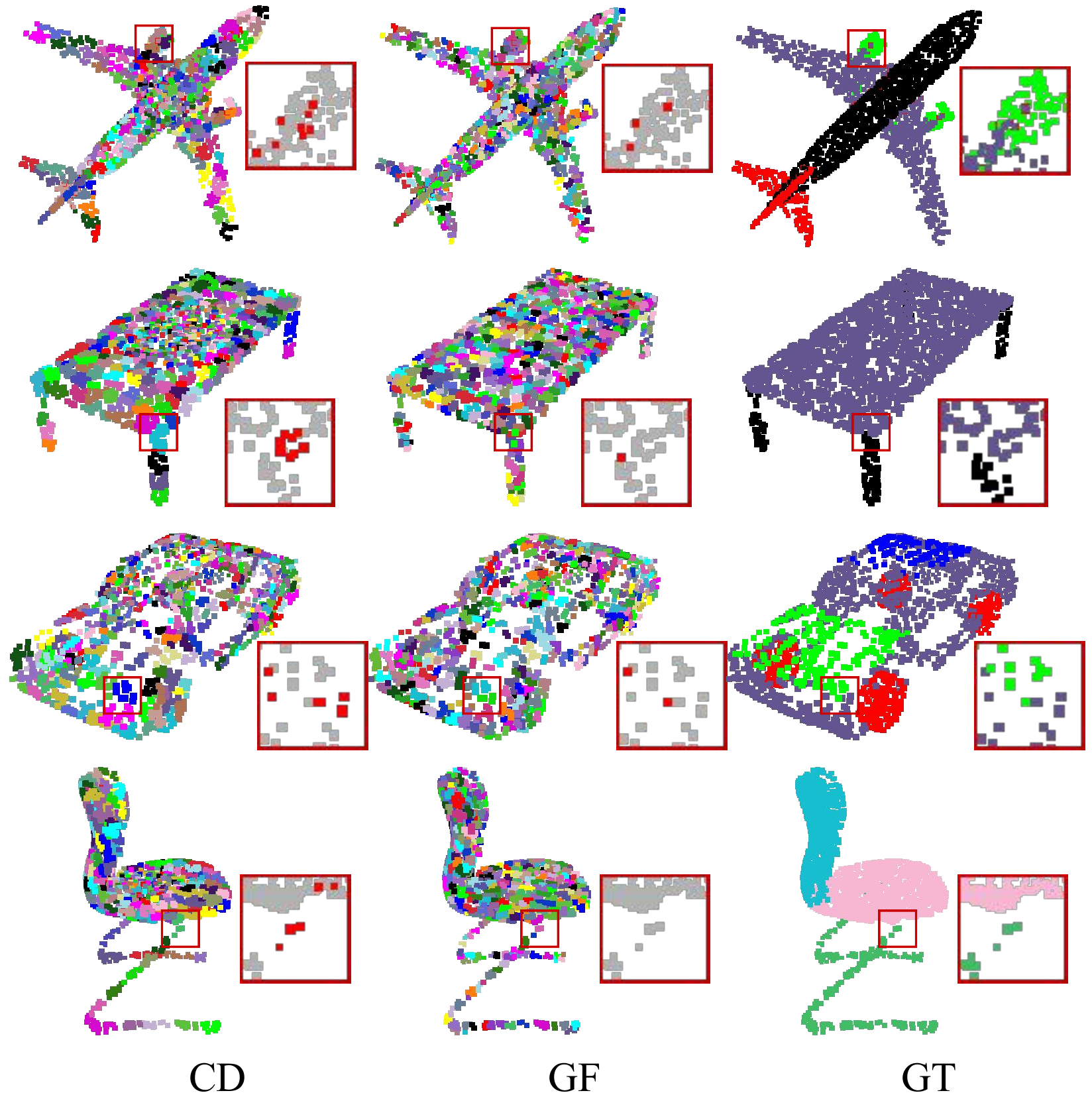}
\caption{Super-points clustered with Coordinate(CD) v.s. with Geometric Features(GF) v.s. Ground Truth(GT). Each shape is clustered into 500 super-points and different clusters are shown in different colors. The red points in magnifying boxes of the first and second columns represent the noise points caused by cross-boundary super-points.}
\label{super-point generation}
\vspace{-0.3cm}
\end{figure}

\begin{table}[!t]
	\caption{Ablation study on ShapeNet with 10k annotations}
	\label{Table Ablation}
	\centering
	\begin{tabular}{c c c c c c}
		\toprule
		SP & Geo     &  Our AL          & Const & DA & mIoU$(\%)$ \\
		\hline
		     -       &     -        &  -           &    -         &      -    & $63.3$ \\
		$\checkmark$ &     -        &  -           &    -         &      -    & $64.4$ \\
		$\checkmark$ & $\checkmark$ &  -           &    -         &      -    & $69.0$ \\
		$\checkmark$ & $\checkmark$ & $\checkmark$ &    -         &      -    & $70.5$ \\
		$\checkmark$ & $\checkmark$ & $\checkmark$ & $\checkmark$ &      -    & $71.4$ \\
		$\checkmark$ & $\checkmark$ & $\checkmark$ & -            & $\checkmark$ & $71.0$ \\
		$\checkmark$ & $\checkmark$ & $\checkmark$ & $\checkmark$ & $\checkmark$ & \bm{$71.8$} \\		
		\bottomrule
	\end{tabular}
\end{table}

\subsection{Additional Study}
{\paragraph{Comparing with other label-efficient learning techniques}
{
With limited annotation budget, some studies have explored point cloud under fewer labeled data. To make fair comparison with existing works, we unified the labeling budget as the cost of labeling a super-point is the same as labeling a point - one manual click needed only. ShapeNet training set consists of 12137 shapes and each shape accounts for 2048 clicks. In Tab.\ref{Limited Labels}, we compare our performances with SONet \cite{2018On}, 3DCNet \cite{zhao20193d}, MUS \cite{9009542}, PTCT \cite{xie2020pointcontrast} and PCWS \cite{xu2020weakly} between $200k$ to $250k$ clicks. The evaluation results and corresponding training label budgets demonstrate that our method performs best among all methods even with the least budget.
}
}

\begin{table}[!t]
	\caption{Comparing different weakly supervised methods on ShapeNet. Clicks indicate the total number of annotation budgets (counted as clicks). Points indicate the total number of points which have labels.}
	\label{Limited Labels}
	\centering
	\resizebox{0.9\linewidth}{!}{
	\begin{tabular}{c c c c}
		\toprule
		Methods & mIoU $(\%)$ & $\#$Clicks & $\#$Points \\	
		\hline
		SONet \cite{2018On} & $64.0$ & $\sim 250k$ & $\sim 250k$ \\
		3DCNet \cite{zhao20193d} & $67.0$ & $\sim 250k$ & $\sim 250k$ \\
		MUS \cite{9009542} & $68.2$ & $\sim 250k$ & $\sim 250k$ \\
		PTCT(HC) \cite{xie2020pointcontrast} & $74.0$ & $\sim 250k$ & $\sim 250k$ \\	
		PTCT(PINCE) \cite{xie2020pointcontrast} & $73.1$ & $\sim 250k$ & $\sim 250k$\\
		PCWS \cite{xu2020weakly} & $74.4$ & $\sim 200k$ & $\sim 200k$ \\
		\hline
		Ours & \bm{$79.9$} & $200k$ & $\sim 710k$ \\
		\bottomrule
	\end{tabular}
	}
\end{table}

\begin{table}[!t]
	\caption{Transfer selected samples to PointNet/PointNet++. Evaluation (mIoU$\%$) is on ShapeNet}
	\label{pointnet/pointnetpp evaluation}
	\centering
	\resizebox{0.9\linewidth}{!}{
	\begin{tabular}{c c c c c c c}
		\toprule
		\multicolumn{2}{c}{\multirow{2}{*}{Full-Supervised}}  & \multicolumn{4}{c}{PointNet(PN)\cite{Qi2017}} & $80.4$ \\
		                                                 & &  \multicolumn{4}{c}{PointNet++(PN++)\cite{qi2017pointnet++}} & $81.9$ \\
		\hline		
		\multicolumn{2}{c}{Annotations Budget} & 10k & 20k & 50k & 100k & 200k \\
		\hline			
		\multirow{3}{*} {PN} & Random & $35.6$ & $44.2$ & $54.3$ & $57.3$ & $65.5$  \\
		& Transfer & \bm{$38.1$} & \bm{$45.6$} & \bm{$54.8$} & \bm{$58.1$} & \bm{$66.1$}  \\
		\hline
		\multirow{2}{*} {PN++} & Random & $36.0$ & $45.8$ & $54.5$ & $57.9$ & $68.9$  \\
		& Transfer & \bm{$39.1$} & \bm{$46.7$} & \bm{$55.5$} & \bm{$59.7$} & \bm{$72.7$}  \\
		\bottomrule
	\end{tabular}
	}
	\vspace{-0.4cm}
\end{table}

\paragraph{Transferability to Other Networks}
{To verify the universal representativeness of the actively selected super-points (Transfer) by DGCNN, we transfer the selected samples to PointNet (PN) and PointNet++ (PN++). As the results shown in Tab.\ref{pointnet/pointnetpp evaluation}, the labeled super-points were selected according to the prediction from DGCNN (transfer), they perform better than random selected super-points (Random) on both PN and PN++. 
}

\vspace{-0.5cm}
\paragraph{Comparisons of Different Super-points}
We examine the quality of super-points generated with and without geometric features by comparing the Noise Rate in the labeled data. As we annotate super-points with a single label, some points would be mislabeled, thus regarded as noise points in training data. 
For comparison, we cluster shape samples of training data into 50/100/200/500/1000 super-points and compare the noise rate with (w/Geo) and without (wo/Geo) geometric features with results reported in Tab.\ref{Noise Rate}. It is evident that at all granularity of super-point, geometric feature will improve the quality of super-point generation by reducing the noise rate.

\vspace{-0.5cm}
\paragraph{Number of Super-Points}
We further evaluate effects of different numbers of super-points clustered in a shape. Given a fixed budget as 10k, we uniformly label super-points and train the network respectively. The evaluations between different super-point data are shown in Tab. \ref{Num of Superpints Comparison}. It is obvious that shapes clustered into 500 super-points achieves the closest to the optimal performance.

\begin{table}[!t]
	\caption{Noise Points Rates $(\%)$ in different clusters. The first line is the number of super-points clustered in one shape.}
	\label{Noise Rate}
	\centering
	\begin{tabular}{c c c c c c}
		\toprule
		\multicolumn{1}{c}{Number of SP} & $50$ & $100$ & $200$ & $500$ & $1000$ \\
		\hline
		Coord & 5.05 & 3.79 & 2.85 & 2.21 & 1.39 \\
		Geo & 4.82 & 3.56 & 2.67 & 1.80 & 1.01   \\
		\bottomrule
	\end{tabular}
\end{table}
\begin{table}[!t]
	\caption{Evaluation with 10k annotations among different numbers of super-points clustered in one shape.}
	\label{Num of Superpints Comparison}
	\centering
	\begin{tabular}{c c c c c c}
		\hline
		Number of SP & $50$ & $100$ & $200$ & $500$ & $1000$ \\
		\hline
		mIoU$(\%)$ & $65.2$ & $67.1$ & $68.0$ & \bm{$69.0$}  & $66.6$     \\
		\bottomrule
	\end{tabular}
\end{table}

\begin{figure*}[!t]
\centering   
\includegraphics[scale=0.5]{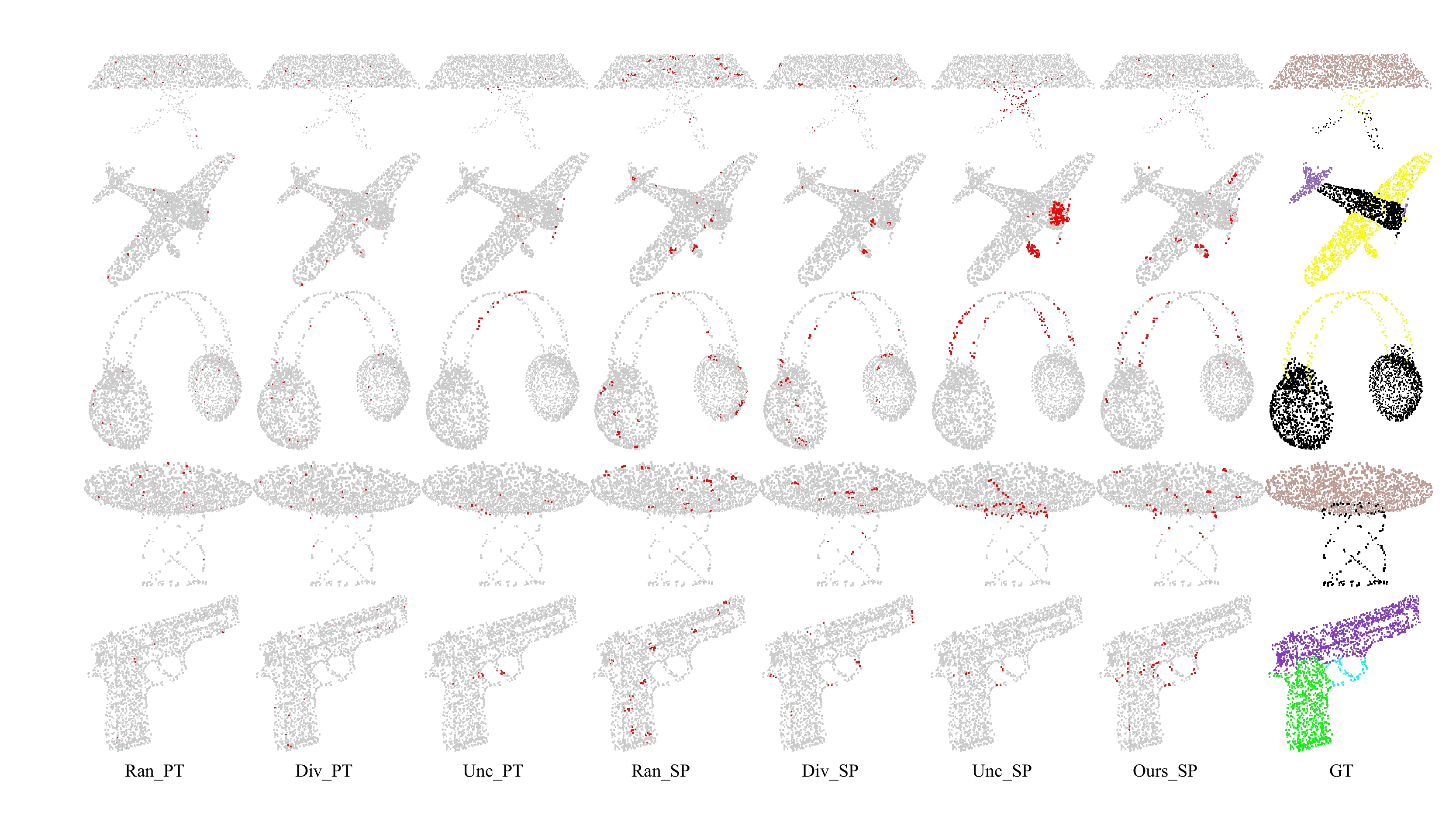}
\caption{We visualize selected points/super-points under different active selection strategies in $200k$ label budget. Ran\_PT, Div\_PT and Unc\_PT respectively represent random, diversity and uncertainty selection on point level data. And Ran\_SP, Div\_SP, Unc\_SP and Ours\_SP are the results of super-point level data selected with random, diversity, uncertainty and our method. The selected units are highlighted in red. In the final column objects semantic segmentation ground truth(GT) are shown.}
\label{selected points and super-points}
\vspace{-0.2cm}
\end{figure*}
\subsection{Qualitative Results}
\paragraph{Visualization of the generated super-points}
{We show super-point examples generated by different methods in Fig.~\ref{super-point generation}. In the first column, super-points are clustered according to their spatial coordinate(CD) only. 
The second column presents the super-points generated with geometric features(GF). The final column is semantic ground truth (GT). In order to show the quality of the different super-points intuitively, we highlight the noise points caused by across boundaries super-points in red in the local enlarged views. It is observed that the number of points across categories can be effectively reduced by adding geometric features.}
\vspace{-0.3cm}
\paragraph{Visualization of the selected points and super-points by different methods}
{To show the distributions of points and super-points selected under different active learning strategies, we highlight the selected points and super-points on several objects in Fig.\ref{selected points and super-points}, where the annotation budget is set as $200k$. In the figure, the first to the third column from left to right respectively represent point level selection under random (Ran\_PT), diversity (Div\_PT) and uncertainty (Unc\_PT) strategies. And the fourth to the seventh column is super-point level selection under random (Ran\_SP), diversity (Div\_SP), uncertainty (Unc\_SP) and our method (Ours\_SP). The selected points/super-points are marked in red. We also show object part semantic segmentation ground truth (GT) in the last column. It can be seen in the picture that randomly selected points/super-points tend to fall in dense areas. Diversity picks points/super-points that are scattered in the feature space, but it cannot distinguish whether these unit samples are difficult to learn for the network. Uncertainty can find unpredictable samples but it is usually too concentrated in a small local region. Our method takes into account the advantages of both diversity and uncertainty, selecting super-points that are more conducive to learning.}